\documentclass[letterpaper, 10 pt, conference]{ieeeconf}
\IEEEoverridecommandlockouts                              

\usepackage{times}
\usepackage{epsfig}
\usepackage{graphicx}
\usepackage{amsmath}
\usepackage{amssymb}
\usepackage{graphicx}
\usepackage{amsmath}
\usepackage{booktabs}
\usepackage{algorithm}
\usepackage{algorithmic}
\usepackage{comment}
\usepackage{graphicx}
\usepackage{amsmath,amssymb} 
\usepackage{color}
\usepackage{booktabs}
\usepackage{bm}
\usepackage{multirow}
\usepackage{tabularx}
\usepackage{array}
\usepackage{nicefrac}       
\usepackage{microtype}      
\usepackage{xcolor}         
\usepackage[utf8]{inputenc} 
\usepackage[T1]{fontenc}    
\usepackage{url}            
\usepackage{booktabs}       
\usepackage{amsfonts}       

\usepackage[pagebackref=true,breaklinks=true,letterpaper=true,colorlinks,bookmarks=false]{hyperref}

\author{Jinchang Zhang*, Ningning Xu*, Hao Zhang, Guoyu Lu
\thanks{* indicates equal contribution. Jinchang Zhang, Ningning Xu, Guoyu Lu are with the University of Georgia. 
        {\tt\small guoyulu62@gmail.com}. Hao Zhang is with University of Massachusetts Amherst.}%
}
\begin{document}

\title{Depth Estimation Based on 3D Gaussian Splatting Siamese Defocus}

\maketitle

\begin{abstract}
Depth estimation is a fundamental task in 3D geometry. While stereo depth estimation can be achieved through triangulation methods, it is not as straightforward for monocular methods, which require the integration of global and local information. The Depth from Defocus (DFD) method utilizes camera lens models and parameters to recover depth information from blurred images and has been proven to perform well. However, these methods rely on All-In-Focus (AIF) images for depth estimation, which is nearly impossible to obtain in real-world applications. To address this issue, we propose a self-supervised framework based on 3D Gaussian splatting and Siamese networks. By learning the blur levels at different focal distances of the same scene in the focal stack, the framework predicts the defocus map and Circle of Confusion (CoC) from a single defocused image, using the defocus map as input to DepthNet for monocular depth estimation. The 3D Gaussian splatting model renders defocused images using the predicted CoC, and the differences between these and the real defocused images provide additional supervision signals for the Siamese Defocus self-supervised network. This framework has been validated on both artificially synthesized and real blurred datasets. Subsequent quantitative and visualization experiments demonstrate that our proposed framework is highly effective as a DFD method.
\end{abstract}

\vspace{-1mm}
\section{Introduction}
\vspace{-1mm}
Depth estimation is crucial for 3D reconstruction and understanding, serving as the foundation for tasks like scene understanding \cite{chen2019towards}, autonomous driving \cite{dong2022towards}, and augmented reality \cite{kalia2019real}. Its success relies on overcoming challenges related to size, speed, accuracy, and cost. Traditional methods use 3D geometric constraints through techniques like structure-from-motion (SfM) \cite{sweeney2015optimizing}\cite{agarwal2011building}\cite{lu2023bird}, image sequences \cite{zhou2017unsupervised}\cite{babu2018undemon}, stereo pairs \cite{garg2016unsupervised}\cite{godard2017unsupervised}\cite{poggi2018learning}, and structured light \cite{ryan2016hyperdepth}. However, monocular SfM faces scale ambiguity, and stereo imaging struggles with calibration and translating disparities into accurate depth. Traditional methods also struggle with defocus blur due to varying focal planes.
\begin{figure*}[t]
\begin{center}
\includegraphics[width=17cm, height=7cm]{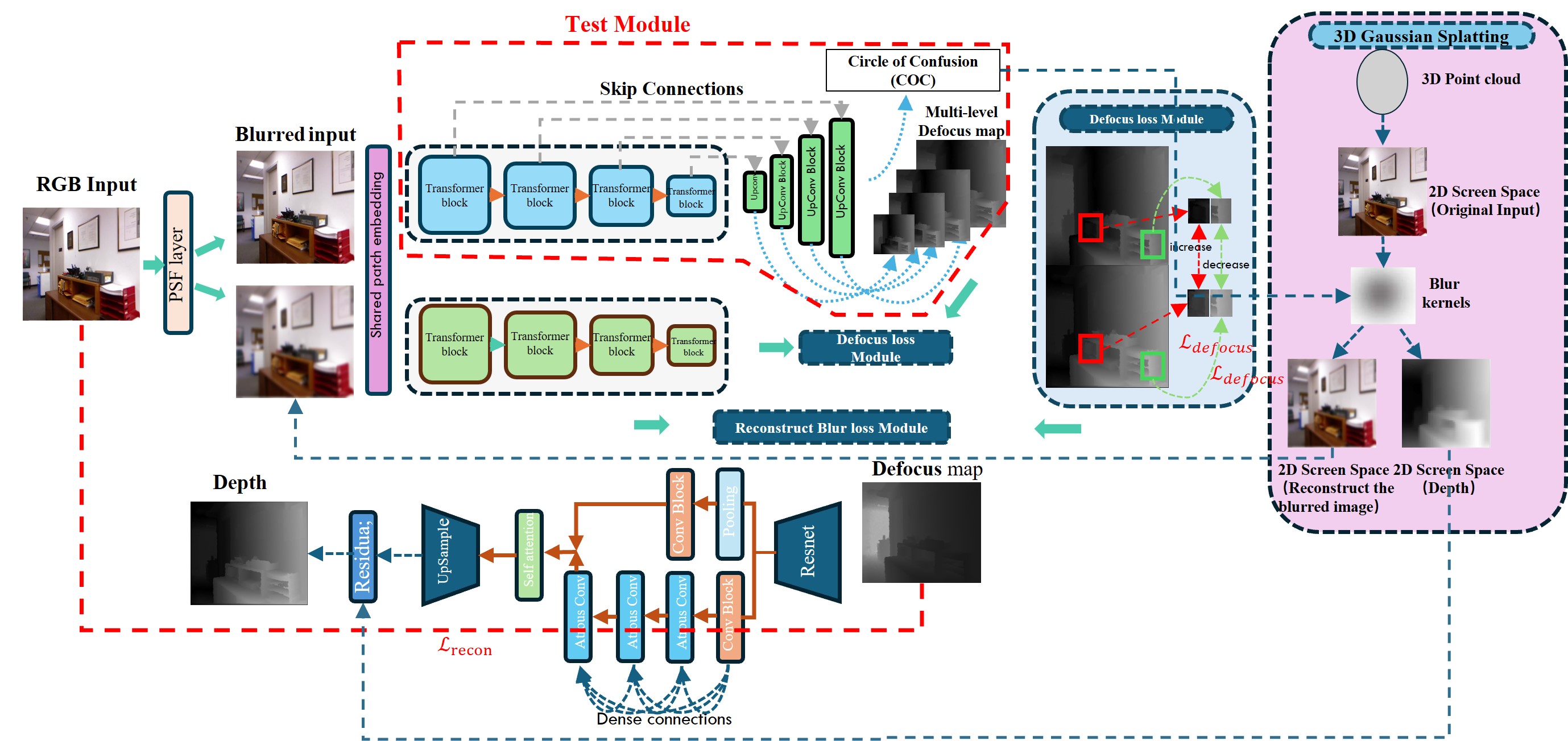}
\end{center}
\vspace{-7mm}
\caption{An overview of the SDNet. We adopt the siamese network structure with mpvit model\cite{lee2022mpvit} and convolutional layer to enhance the defocus map modeling. We use defocus loss module\cite{tao2023siamese} to learn the relationship between distance and blurriness while training the Siamese network. After training, we use one single blurred image to predict the defocus map for the depth inference.}
\vspace{-7 mm}
\label{fig:network}
\end{figure*}
Depth from Defocus (DFD) is an alternative approach to depth recovery that relies on defocus blur. Unlike traditional Structure-from-Motion (SfM), DFD estimates depth by utilizing the geometry of the camera lens and depth variations. Previous research \cite{suwajanakorn2015depth} has shown that generating a focal stack of images with different levels of blur and analyzing the blur in each image can provide depth information at various focal distances. Supervised and self-supervised DFD methods based on deep learning have since been developed to estimate depth using focal stacks and All-In-Focus (AIF) images. However, current DFD methods that use AIF images often rely on approximations, treating small-aperture photos as AIF images, which can lead to inaccuracies. Additionally, using focal stacks in real-world scenarios is impractical due to the need for frequent focus adjustments. To overcome these limitations, this paper introduces Siamese-Defocus-Net (SDNet), a neural network capable of simultaneously estimating defocus information and depth from a single image, eliminating the need for multiple images or focus adjustments and simplifying depth estimation in dynamic environments.

To estimate depth from a single defocused image, we begin with the camera lens model and focus on estimating the defocus map, a crucial step in depth recovery. This paper introduces a defocus depth estimation method, trained on focal stacks but designed to estimate depth from a single defocused image during testing. We develop a self-supervised framework combining the Siamese Defocus Network and 3D Gaussian splatting, training both models jointly.
The Siamese Defocus Network takes the same image with varying levels of defocus as input and outputs the corresponding defocus map and Circle of Confusion (CoC) for each image. To ensure accurate prediction of defocus maps across different levels of blur, we leverage Siamese networks and train the model using defocus loss. This approach allows the network to effectively capture defocus characteristics from the focal stack, validating its ability to distinguish varying blur levels and improving its sensitivity to defocus features. This training strategy enhances the network's performance when processing images with different focal lengths, and it can extract and integrate features across multiple scales, improving the accuracy of defocus map generation.
To further improve CoC and defocus map predictions, the CoC is fed into the 3D Gaussian splatting model to verify its accuracy. Specifically, we input a series of defocused images into the 3D Gaussian splatting model, combining the CoC predicted by the Siamese Defocus Network with 2D projection images generated by the splatting model. We render defocused images and compute the differences between synthetic and real defocused images, providing additional supervision to the Siamese Defocus Network. This approach explores the connection between defocus characteristics and the camera lens model, enabling effective depth estimation. We validated our method across multiple datasets using the point spread function (PSF), rendering images with varying focal lengths and demonstrating the feasibility and effectiveness of our approach. The framework is illustrated in Figure \ref{fig:network}.

In summary, the contributions of this paper include:
1. We propose a system that can simultaneously estimate defocus maps and scene depth.
2. We design a self-supervised framework based on Siamese networks and 3D Gaussian splatting, capable of generating defocus maps and Circle of Confusion (CoC) at different focal lengths.
3. we embed 3D Gaussian splatting into the Siamese Defocus Network, using CoC as input to the 3D Gaussian splatting to calculate the blur reconstruction loss, thereby improving the training of the Siamese Defocus Network. 3D Gaussian splatting generates the initial depth, which, combined with the predicted defocus map, serves as input for the depth estimation network. The depth estimation network uses the defocus map to optimize the initial depth and predict depth residuals.

\vspace{-2mm}

\section{Related Work}
\textbf{Monocular depth estimation.} Monocular depth estimation aims to reconstruct depth information from a single camera by utilizing scene geometry as a training constraint. \cite{godard2019digging}\cite{lu2023deep}\cite{lu2024slam} employ consecutive images, using both depth and pose networks to predict depth. Subsequent works \cite{lyu2021hr,he2022ra} have successfully applied monocular depth estimation algorithms to high-resolution images, achieving impressive results. In contrast, stereo depth estimation reconstructs depth using the baseline between two cameras and the disparity map, avoiding issues with scale ambiguities. \cite{tonioni2019real} and \cite{xu2022attention} calculate 3D and 4D cost volumes to recover depth information. 
By using the camera model to calculate road depth, \cite{zhang2024embodiment} provides depth priors for depth estimation.
However, due to significant information loss in the encoded volume, these methods do not achieve optimal accuracy.

\textbf{Defocus map and depth from defocus.}
A defocus map quantifies the level of defocus blur or the size of the circle of confusion (CoC) for each pixel in a blurred image. For estimating defocus maps, 
Depth from defocus (DFD) methods derive depth by measuring image blurriness caused by lens effects. 
\cite{zhou2009coded} made advancements by applying coded aperture cameras to measure defocus blur.
\cite{gur2019single} predicts depth by simulating defocus on datasets like KITTI and Make3D, while \cite{maximov2020focus} reconstructs both all-in-focus images and depth using supervised learning. \cite{yang2022deep} improves depth estimation by incorporating focus volume and differential focus volume into their model, enhancing accuracy. \cite{si2023fully} introduces a fully self-supervised framework that estimates depth from a sparse focal stack. These deep learning approaches rely on focal stacks or all-in-focus (AIF) images, using multiple images with varying degrees of blurriness for depth estimation. However, this is impractical in real-world applications. To address these limitations, we leverage prior knowledge from the camera lens model and extract multi-scale features from a single blurred image, enabling depth estimation without requiring multiple images.
\vspace{-1mm}
\section{Depth from Defocus Framework}
In this section, we first introduce the foundational background knowledge on the camera lens model and the Depth from Defocus (DFD) method, followed by a detailed presentation of our network architecture for defocus and depth estimation. We developed a Siamese network architecture to estimate defocus maps and the Circle of Confusion (CoC) from two differently blurred images of the same scene. 3D Gaussian splatting is integrated to render blurred images based on the predicted CoC, and blur reconstruction loss is applied to optimize the CoC predictions. The final depth is then estimated using the predicted defocus maps and the initial depth information obtained from 3D Gaussian splatting.
\subsection{Camera Lens Model}
\begin{figure}[t]
	\centering
	\includegraphics[width=1.0\linewidth]{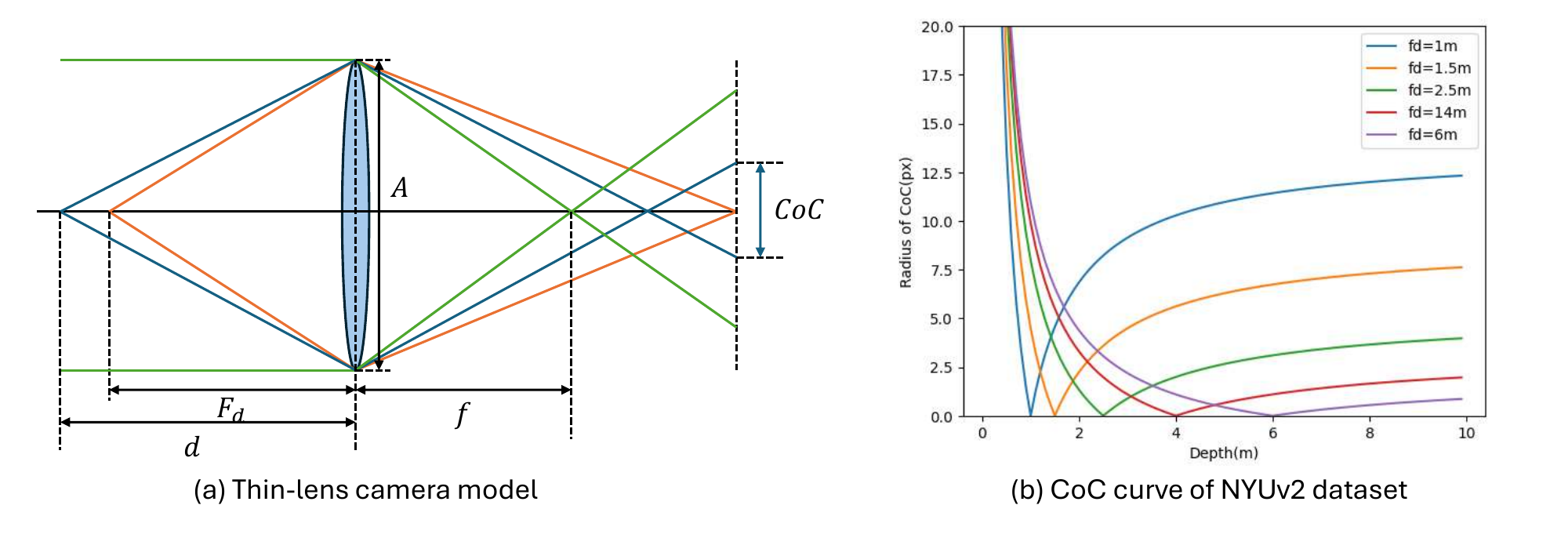}
 \vspace{-10mm}
	\caption{(a) An illustration of the camera Thin-Lens model. Objects on the focal plane (indicated by the orange line) are sharply imaged, while objects off the focal plane appear blurred due to the Circle of Confusion (CoC).
(b) The CoC curve derived from the NYUv2 dataset demonstrates the relationship between object depth and the blur radius, where the blur radius initially decreases as depth increases and then enlarges.}
	\label{fig:camera}
 \vspace{-6mm}
\end{figure}

Depth from defocus methods are based on thin lens model and geometry properties, as shown in Fig. \ref{fig:camera} (a), which explains the origin of blurriness generation. When an object is situated on the focal plane, any given point on the object corresponds to a specific point on the image plane. However, when the object is moved away from the focal plane, each point on the object corresponds to a circular region with a radius of sigma on the image plane, thereby resulting in blurriness. This phenomenon is referred to as the 'Circle of Confusion' (CoC). Based on the Gaussian lens formula and the properties of similar triangles, we can derive the equation:
\vspace{-2mm}
\begin{equation}
    CoC = A\frac{|d-F_d|}{d}\frac{f}{F_d-f},
    \vspace{-1mm}
\end{equation}

\noindent where $A$ is the diameter of the lens, $F_d$ is the focus distance, $f$ is the focal length, and $d$ is the distance of the object to the lens (depth). In general, the unit of depth is meters, so we must use the CMOS pixel size $p$ to convert the unit of CoC to meters as well. For convenience, we incorporate the camera parameter f-number $N$, defined as $f/A$, into the formula, and $\sigma$ is the radius of CoC. This gives the following equation:
\vspace{-2mm}
\begin{equation}
    \sigma=\frac{|d-F_d|}{d}\frac{f^2}{N(F_d-f)}\frac{1}{2p}\label{coc formula}
    \vspace{-1mm}
\end{equation}

Figure \ref{fig:camera} (b) reveals that CoC sharply decreases as the subject distance approaches the focus distance, denoting a clear boundary between in-focus and out-of-focus regions. Beyond the focus distance, the CoC incrementally increases, indicating a gradual onset of blur with depth.

\subsection{Defocus Generation from Camera Lens Model}
\label{Defocus Generation}
As mentioned in the previous section, we can generate defocus blurred images from equation (\ref*{coc formula}) by simulating the CoC. The most common used method for generating defocus blurred image is PSF method. The point spread function describes the response of the camera lens model to a point source or point object. We employ a Gaussian kernel function to generate synthetic training images, following the methodologies of previous work\cite{si2023fully,gur2019single}. 
\vspace{-2mm}
\begin{equation}
\vspace{-2mm}
  G_{x,y}(u,v)=\frac{1}{2\pi\Sigma_{x,y}^2}exp(-\frac{u^2+v^2}{2\Sigma_{x,y}^2}),\label{gaussian}
\end{equation}

\noindent where $\Sigma_{x,y}$ represents the defocus map, which is generated based on depth and the camera optical model. Let $I_A$ denote the all-in-focus image, and $I_{defocus}$ represent the image generated with defocus blur. In practice, the $I_A$ is very difficult to obtain. Usually, we acquire an image $I_R$ with some depth of field, which already contains some blur. We can assume that the $I_R$ can be represented as the $I_A$ blurred by an initial defocus blur kernel $G_0$. 
The blurred image is generated by convolving the all-in-focus image with the Gaussian kernel, as shown in the following equation:
\vspace{-2mm}
\begin{equation}
\begin{aligned}
\vspace{-3mm}
I_{defocus} &= I_A \otimes G, I_R = I_A \otimes G_0, \\
I_{defocus} &= I_R \otimes G  = I_A \otimes (G_0 \otimes G).
\end{aligned}
\end{equation}

We can represent the $I_R$ as a blurred version of the $I_A$. Although the $I_A$ is the most fundamental image in an ideal case, it is difficult to obtain in practice. The actual image $I_R$ already contains some level of blur. Therefore, we can directly use  $I_R$ as input and apply additional blur $G$ to generate a stronger defocus-blurred image.
Here, $G$ is the spatially varying Gaussian kernel, which changes with pixel positions $x, y$ to reflect varying degrees of defocus. The blur kernel $G_0$ represents the initial level of blur in the image $I_R$, which may be determined by factors such as the optical properties of the camera and the depth of field. Since the defocus map $\Sigma_{x,y}$ is pixel-dependent, the convolution kernel $G$ also varies spatially. We directly adopt the PSF layer proposed in \cite{gur2019single} to generate the blurred images. The window size of the convolution kernel is set to 7, and we apply thresholding to the blur radius as $\sigma = \sigma \cdot \mathbf{1}_{\sigma \geq 1}$ to avoid negligible blur effects from small radii.



\subsection{Siamese Defocus Net for Defocus Mapping}
The Circle of Confusion (CoC) is the core principle behind the phenomenon of defocus. Based on CoC mapping, we built a self-supervised framework that combines Siamese Defocus Net and 3D Gaussian splatting to accurately estimate defocus maps and CoC. As described in Section \ref{Defocus Generation}, we generated defocused images at different focal lengths from the camera lens model, forming a focal stack. To enhance Siamese Defocus Net's learning of defocus maps, we adopted a Siamese network structure that takes images with varying focal distances as input. To optimize the prediction of CoC and defocus maps, we integrated CoC into the 3D Gaussian splatting model to generate synthetic blurred images. We then compared the synthetic images with real defocused images, using the differences for self-supervised training to optimize network parameters and improve prediction accuracy.

Leveraging our Siamese network design, we effectively learn varying degrees of blurriness in the same region at different focus distances. Specifically, we incorporate the multi-path transformer from \cite{lee2022mpvit}, which allows for both fine and coarse feature representations. The network uses a multi-scale patch embedding strategy through overlapping convolutions, processing these embeddings in parallel paths within the Transformer framework. This enables independent and efficient handling of multi-scale features, capturing both detailed and broader contextual information for dense prediction tasks.
For predicting defocus maps, it is essential to extract both local and global features to understand the blur characteristics associated with different focus distances. After extracting local features from each patch, we apply a max-pooling operation to fuse local and global features, allowing the network to accurately capture the blur level for each focus distance. This approach is inspired by the layer-wise global pooling concept from \cite{maximov2020focus}. During training, defocus loss is applied to paired patches to measure the similarity between corresponding patches, effectively capturing the variations in defocus blur across different focus distances.
SDNet plays a vital role in our pipeline by accurately predicting the defocus map of a given input blurred image. This defocus map is integral to our methodology, as it serves as the foundation for reconstructing the original blurred input image to be sharper and clearer. Importantly, we utilize the defocus map to derive the depth of the image, which is essential for understanding the spatial relationships and focus levels within the scene. Therefore, the precision of our defocus map prediction directly impacts the overall effectiveness and reliability of the image reconstruction process. Here, we employ not only a reconstruction loss but also a loss that assesses the degree of image blurring, as well as a smoothness loss.
We first introduce defocus loss, following the work\cite{tao2023siamese}. The defocus loss is applied on image patches to learn defocus within each focal stack. The smoothness loss is utilized to ensure that the gradients of the target image are not excessively large, thus enhancing smoother gradient changes. Let $D_1, D_2$ represent the predicted defocus map of the SDNet. the loss constraints are:
\vspace{-2.5mm}
\begin{equation}
\vspace{-2mm}
    \mathcal{L}_{defocus}=\mathbf{E}\left[-cosine(D_1, D_2) \right]
\end{equation}
We utilize the Laplacian operator to manipulate the edge map and compute its variance. For the prediction of the defocus map, we aim to exhibit distinct boundaries between blurred and non-blurred regions, which forces large gradient changes. Following the work in \cite{lu2021self}, we take the negative logarithm, sum it up, and normalize the result, which is expressed as:
\vspace{-1mm}
\begin{equation}
\vspace{-2mm}
    \mathcal{L}_{blur}=-\frac{1}{N}\sum\beta log\left(\frac{\sum_i\sum_j (\nabla^2\hat{I}(i,j))^2}{M-\mu^2}\right)
\end{equation}

\noindent where $\nabla^2$ is the Laplacian operator, $M$ is the amount of pixels, $\mu$ is the mean value of pixels, and $beta$ is a scaling factor, which we have set to 0.01 for the subsequent experiments.

\subsection{ 3D Gaussian Splatting Siamese}
3D Gaussian Splatting \cite{kerbl20233d} is a technique used in point cloud data processing. Consider a 3D point cloud 
$\mathbf{P} = \{ \mathbf{p}_i \in \mathbb{R}^3 \mid i = 1, 2, \ldots, N \}$, where $\mathbf{p}_i$ represents a point in the point cloud. 3D Gaussian Splatting can be expressed as applying the following operation to each point ${G_i}\left( {\rm P} \right)$:
\vspace{-2mm}
\begin{equation}
\vspace{-2mm}
   {G_i}\left( {\rm P} \right) = \exp \left( { - \frac{1}{2}{{\left( {{\rm P} - {\mu _i}} \right)}^T}\Sigma_i^{-1}\left( {{\rm P} - {\mu _i}} \right)} \right)
\end{equation}


Specifically, each Gaussian function is defined by the following attributes: a center position \(u\), a covariance matrix \(\Sigma\) derived from anisotropic scaling \(s\) and a quaternion vector \(q\), as well as opacity \(o\) and spherical harmonics coefficients \(h\). To evaluate the accuracy of the coc calculation, we need to project the 3D Gaussian points into 2D screen space. The 2D Gaussian in screen space is formulated as:
\vspace{-3mm}
\begin{equation}
\vspace{-2mm}
    {G_i}^\prime \left( x \right) = \exp \left( { - \frac{1}{2}{{\left( {x - {{\mu '}_i}} \right)}^T}{{\left( {{\sum _i}^\prime } \right)}^{ - 1}}\left( {x - {{\mu '}_i}} \right)} \right)
\end{equation}
\begin{equation}
\vspace{-2mm}
    {\sum _i}^\prime  = JW{\sum _i}{W^T}{J^T}
\end{equation}
The Jacobian matrix J of the projective transformation can be computed during the process of projecting the 3D Gaussian points onto 2D screen space.W represents the view matrix transforming points from world space to camera space. where \( u'_i \) represents the 2D center position, post-projection.

To help Siamese Defocus Net better predict the defocus map and the Circle of Confusion (CoC), we incorporate a depth-of-field rendering process into the 3D Gaussian splatting. We begin by estimating the camera pose and the initial sparse point cloud from the defocused image. For each viewpoint, we introduce the Circle of Confusion predicted by the Siamese Defocus Net for the blurred image. During the optimization process, for each sampled viewpoint, we render a defocused image using the CoC to fit the target view.
We hypothesize that a blurred image can be obtained by first applying depth-dependent blurring to individual Gaussians in the scene and then rendering the image from these blurred Gaussians \cite{krivanek2003fast}. As illustrated in Figure \ref{fig:network}, during the blur rendering process, the 3D Gaussian functions \(G_i\) are projected onto the 2D screen space. Each 2D Gaussian \(G'_k\) (representing the projection of a 3D Gaussian point) is convolved with a Gaussian blur kernel proportional to the CoC generated by a camera lens model with a finite aperture, as shown in Figure \ref{fig:camera} (left). The adopted camera lens model includes an aperture parameter \(Q\), which differs from the pinhole model. Object points that deviate from the focal distance \(f\) form a region known as the CoC, rather than a single point. The final color is obtained by compositing the convolved Gaussians \(G''_i\). We assume that the depth across each 2D Gaussian's support region is uniform, set as \(z_k\), which is the z-coordinate of the transformed center position in camera space. Based on the CoC radius $\sigma $ calculated using Equation \ref{coc formula}, we construct the blur kernel $g_i^\sigma  = 1/2\exp \left( {{x^T}\Sigma _i^\sigma x} \right)$ 
When $\Sigma _i^\sigma  = aI,a = \frac{1}{{2\ln 4}}\left( {{\sigma _i}} \right)$ \cite{wang2024dof}, $g_i^\sigma$ is similar to a uniform intensity distribution within the CoC. Using Gaussian kernels for blur convolution ensures that the subsequent color composition remains similar to the original rasterization process, as convolving two Gaussians yields another Gaussian. The convolved 2D Gaussian is defined as \(G''_i = G'_i \ast g_i^\sigma\), where \(\ast\) denotes convolution.
Although \( G''_i \) has an infinite support in theory, in practice it is truncated by a cutoff radius \( t \) and is evaluated only for a limited range. Therefore, each pixel \( x \) is just associated with a part of the Gaussians within the scene, whose number is denoted as \( N_x \). Finally, the projected Gaussians are rendered through alpha blending:
\vspace{-2mm}
\begin{equation}
\vspace{-2mm}
    \hat I\left( x \right) = \sum\limits_{i = 1}^{{N_x}} {{T_i}{\alpha _i}{c_i}} ,{\alpha _i} = {G''_i} \left( x \right),{T_i} = \prod\nolimits_{j = 1}^{i - 1} {\left( {1 - {\alpha _j}} \right)} 
\label{ix}
\end{equation}
\({T_i}\) is transmittance, \(c_i\) denotes the view-dependent color of the i-th Gaussian associated with the queried pixel \(x\). As Equation \ref{ix} is fully differentiable, 3DGS reconstructs a 3D scene by minimizing errors between its renderings and training views.

\vspace{-3mm}
\begin{equation}
\vspace{-4mm}
\label{3dgausspere}
\mathop {\min }\limits_{\left\{ {S} \right\}} \sum\limits_{m - 1}^M {\mathcal{L}_{rec}\left( {{{\widehat I}_m},{I_m}} \right)} 
\end{equation}

\begin{equation}
\vspace{-1mm}
    \mathcal{L}_{rec} =  \left( \alpha \frac{1 - SSIM(\hat I, I)}{2} + (1 - \alpha) \|\hat I - I\|_1 \right)
\end{equation}
 As shown in Equation \ref{3dgausspere}, the optimizable parameters now include the underlying 3D scene \(S\) and the CoC parameters \(\{M_m\}_{m=1}\) for the training views.
 m is the index running over the training views. $I$is the real blurred image, and ${\hat I}$ is the CoC-synthesized blurred image.
 The 3D Gaussian splatting module generates a blurred image $\hat I\left( x \right)$ along with its corresponding depth $\hat D\left( x \right)$, defined as:
 \vspace{-2mm}
 \begin{small}
\begin{equation}
\vspace{-1mm}
    \hat D\left( x \right) = \sum\limits_{i = 1}^{{N_x}} {{T_i}{\alpha _i}{z_i}} ,{\alpha _i} = {G''_i} \left( x \right),{T_i} = \prod\nolimits_{j = 1}^{i - 1} {\left( {1 - {\alpha _j}} \right)} 
\label{dx}
\end{equation}
\end{small}
By introducing the defocus map and CoC calculated by the Siamese Defocus network, the scale of the depth obtained through 3D Gaussian splatting is made closer to the true depth. However, 3D Gaussian splatting relies on sparse point cloud data, and the generated depth may have some inaccuracies. The limitations of the sparse point cloud data itself can lead to incomplete depth information, and errors introduced during projection and rendering are also non-negligible. Therefore, it is necessary to further refine and optimize the depth.

\vspace{-2mm}
\subsection{Joint Optimization}
We simultaneously trained the 3D Gaussian Splatting model and Siamese Defocus Net. As described in Section \ref{Defocus Generation}, we generated blurred images using the given CoC parameters. For Siamese Defocus Net, we input images with varying degrees of blur to predict the CoC and defocus map for each image. We selected a set of scenes with the same level of blur as input for COLMAP to estimate the camera poses and initial sparse point cloud. The output from COLMAP was used as the initialization for the 3D Gaussian Splatting model. During the optimization process, we rendered blurred images using the CoC parameters predicted by Siamese Defocus Net and compared them to the original blurred images to assess whether the reconstructed 3D scene could accurately reproduce the input blur effects. At the same time, this process helped Siamese Defocus Net further refine its CoC predictions. Then, we obtained the total loss as follows:
\vspace{-2mm}
\begin{equation}
\vspace{-1mm}
    \mathcal{L}=\mu_1\mathcal{L}_{defocus} + \mu_2\mathcal{L}_{blur}+\mu_3\mathcal{L}_{recon}
\end{equation}

\subsection{Defocus-based Depth Estimation}
This paper introduces DepthNet, designed to improve depth estimation accuracy by utilizing the defocus map generated by DefocusNet and initial depth data from 3D Gaussian Splatting. DepthNet enhances accuracy by predicting the residual between the estimated depth and the ground truth.
Using an encoder-decoder structure, the encoder is built on ResNet, enhanced by integrating an Atrous Spatial Pyramid Pooling (ASPP) module \cite{chen2017deeplab}. This setup leverages ResNet's feature extraction strengths, while the ASPP module improves the network's ability to capture multi-scale contextual information, essential for accurate depth estimation.
We further propose an improved ASPP module with dense connections between the 1x1 convolution and Atrous convolution layers, promoting better feature integration to capture both local and global details. A self-attention mechanism is also included to refine feature extraction. The decoding process incorporates three upsampling blocks, with skip connections to maintain high-resolution details, ensuring accurate defocus depth prediction. DepthNet is trained using L1 loss and smoothness loss \cite{godard2017unsupervised}, estimating the residual between the 3D Gaussian Splatting depth and the ground truth.
\vspace{-2mm}
\begin{equation}
\vspace{-1mm}
    \mathcal{L}_{res} = \frac{1}{N} \sum^N \left| (\hat{D}_{x} +  {D}_{res}) - D_{gt} \right|_1
\end{equation}
\vspace{-3mm}
\begin{equation}
\vspace{-2mm}
    \mathcal{L}_{sm} = \frac{1}{N} \sum^N \left( |\partial_x \hat{D}| e^{-|\partial_x I|} + |\partial_y \hat{D}| e^{-|\partial_y I|} \right)
\end{equation}
\(\hat{D}_{x}\) is the depth obtained from 3D Gaussian splatting, \(D_{res}\) is the DepthNet result, and \(D_{gt}\) is the ground truth. 
\vspace{-2mm}
\section{Experiment}
We present the quantitative and visual results of our experiments as well as the ablation study. Our experiments are conducted on both synthetically generated datasets and real defocus datasets.

\subsection{Implementation Details}
\textbf{Synthetic dataset:}. The FoD500 dataset \cite{maximov2020focus}  contains 1000 scenes, each focal stack comprising 5 RGB images, 5 defocus maps, 1 depth map, and 1 all-in-focus image. The dataset is set with a max distance of 3 meters, and focus distances are defined at 0.3, 0.45, 0.75, 1.2, 1.8 meters.
\textbf{Synthetic Defocus with Real Images}: The synthetic dataset is generated using the method described in Section \ref{Defocus Generation}. We used the NYUv2 indoor dataset \cite{Silberman:ECCV12} for our experiments. The NYUv2 dataset is set with a maximum depth limit of 10 meters. For NYUv2, we set the focus distances at [1, 1.5, 2.5, 4, 6] meters to generate defocus blur.
\textbf{Real Focal Stack Dataset}: The MobileDFF dataset \cite{suwajanakorn2015depth} contains 11 scenes, with the number of focal stacks per scene ranging from 14 to 33.
\vspace{-2mm}
\subsection{Comparison with State-of-the-art methods}

\renewcommand{\arraystretch}{1.2} 
\vspace{-2mm}
\begin{table}[h]
\begin{center}
\resizebox{0.49\textwidth}{!}{
\begin{tabular}{lcccccc}
\toprule
Methods & $\delta_1$ & $\delta_2$ & $\delta_3$ & $RMSE$ & $AbsRel$ \\ \midrule
\multicolumn{6}{c}{Regular} \\ \midrule
DefocusNet\cite{lu2021self} & 0.912 & 0.967 & 0.983 & 0.194 & 0.090 \\
DFF-FV\cite{yang2022deep} & 0.883 & 0.953 & 0.980 & 0.231 & 0.107 \\
DFF-DFV\cite{yang2022deep} & 0.921 & 0.977 & 0.990 & 0.219 & 0.104 \\
DIAF-net\cite{si2023fully} & 0.746 & 0.883 & 0.938 & 0.351 & 0.177 \\
{Ours} & {0.849} & {0.930} & {0.983} & {0.256} & {0.173} \\ \midrule
\multicolumn{6}{c}{0.5m} \\ \midrule
DefocusNet\cite{lu2021self} & 0.911 & 0.933 & 0.938 & 0.062 & 0.069 \\
DFF-FV\cite{yang2022deep} & 0.977 & 0.996 & 0.999 & 0.023 & 0.032 \\
DFF-DFV\cite{yang2022deep} & 0.976 & 0.996 & 0.999 & 0.023 & 0.031 \\
DIAF-net\cite{si2023fully} & 0.889 & 0.987 & 0.992 & 0.072 & 0.138 \\
{Ours} & {0.930} & {0.990} & {0.996} & {0.057} & {0.079} \\ \bottomrule
\end{tabular}}
\end{center}
\vspace{-3mm}
\caption{The quantitative depth comparison of the FoD500 dataset.}
\label{Tab:FoD}
\vspace{-7mm}
\end{table}

\textbf{FoD500 Dataset} Following the previous DFD work \cite{si2023fully}, we conducted evaluations using two data splits: regular (including all results) and for depth less than 0.5 meters. We split the 500 focal stacks into 400 training stacks and 100 testing stacks. For comparison, we selected currently open-sourced DFD methods (DAIFNet \cite{si2023fully}, AiFDepthNet \cite{wang2021bridging}, DFF-DFV/FV \cite{yang2022deep}) and presented the quantitative comparison results in Table \ref{Tab:FoD}. Our approach takes only a single defocus image as input, whereas DFF-DF \cite{yang2022deep}, DFF-DFV \cite{yang2022deep}, and DAIF-net \cite{si2023fully} use focal stacks (5 images) as input. Our model, using only a single defocus image, achieves results comparable to pther methods that use focal stacks as input. 

\begin{table}[h]
\begin{center}
\resizebox{0.49\textwidth}{!}{
\begin{tabular}{l|l|ccc|cc}
\toprule
		Methods     & Input       & $\delta_1\uparrow$ & $\delta_2\uparrow$ & $\delta_3\uparrow$ & $RMSE\downarrow$ & $AbsRel\downarrow$ \\
		\midrule
		Moeller\cite{moeller2015variational}     & focal stack  & 0.670 & 0.778 & 0.912 & 0.985 & 0.263 \\
		Suwajanakorn\cite{suwajanakorn2015depth} & focal stack  & 0.688 & 0.802 & 0.917 & 0.950 & 0.250 \\      
		Gur and Wolf\cite{gur2019single}         & in-focus     & 0.720 & 0.887 & 0.951 & 0.649 & 0.184 \\
		Defocus-Net\cite{lu2021self}             & defocus      & 0.732 & 0.887 & 0.951 & 0.623 & 0.176 \\
		Focus-Net\cite{lu2021self}               & focal stack  & 0.748 & 0.892 & 0.949 & 0.611 & 0.172 \\
		AiFDepth-Net\cite{Wang-ICCV-2021}        & focal stack  & 0.688 & 0.944 & 0.961 & 0.669 & 0.289 \\
		DAIF-Net\cite{si2023fully}               & focal stack  & 0.950 & 0.979 & 0.987 & 0.325 & 0.170 \\
		DFF-FV\cite{yang2022deep}                & focal stack  & 0.956 & 0.979 & 0.988 & 0.285 & 0.470 \\
		DFF-DFV\cite{yang2022deep}               & focal stack  & 0.967 & 0.980 & 0.990 & 0.235 & 0.445 \\
		\textbf{Ours}       & \textbf{defocus}      & \textbf{0.964} &\textbf{0.998} & \textbf{0.999} & \textbf{0.201} & \textbf{0.026} \\
\bottomrule
\end{tabular}}
\end{center}
\vspace{-3mm}
\caption{The quantitative depth comparison of the NYUv2 dataset. }
\label{Tab:Comparison}
\vspace{-10mm}
\end{table}

\textbf{NYUv2 Dataset:} In the Depth from Defocus task on the NYUv2 dataset, we utilized the focal stack generation technique from \cite{si2023fully} and compared our model with two DFD methods and four self-supervised deep learning methods. Using the provided code for DAIF-net and AIFDepth-net, we evaluated their results. For other focal stack-based methods, where source code was unavailable, we directly referenced the results provided in \cite{si2023fully} for comparison. As shown in Table \ref{Tab:Comparison}, despite using only a single defocused image as input, our method performs comparably to current state-of-the-art methods on key metrics. Additionally, we compared our approach with several monocular depth estimation methods. As demonstrated in Table \ref{Tab:NYU other}, our method consistently outperforms the current SOTA monocular depth estimation methods, highlighting the broad applicability of this technique.

\begin{table}[h]
\vspace{-2mm}
\begin{center}
\resizebox{0.49\textwidth}{!}{
\begin{tabular}{@{}lccccc@{}}
  \toprule
  Methods & $RMSE\downarrow$ & $AbsRel\downarrow$ & $\delta_1\uparrow$ & $\delta_2\uparrow$ & $\delta_3\uparrow$ \\ \midrule
  SharpNet\cite{ramamonjisoa2019sharpnet} & 0.502 & 0.139 & 0.836 & 0.966 & 0.993 \\
  AdaBins\cite{bhat2021adabins} & 0.364 & 0.103 & 0.903 & 0.984 & 0.997 \\
  NeWCRFs\cite{yuan2022new} & 0.334 & 0.041 & 0.922 & 0.992 & 0.998 \\
  ZoeDepth\cite{bhat2023zoedepth} & 0.270 & 0.032 & 0.955 & 0.995 & 0.999 \\
  EVP\cite{lavreniuk2023evp} & 0.224 & 0.027 & 0.976 & 0.997 & 0.999 \\ \midrule
  \textbf{Ours} & \textbf{0.201} & \textbf{0.026} & \textbf{0.964} & \textbf{0.998} & \textbf{0.999} \\ \bottomrule
\end{tabular}}
\end{center}
\vspace{-3mm}
\caption{Comparison with monocular depth estimation on the NYUv2.}
\label{Tab:NYU other}
\vspace{-7mm}
\end{table}

\begin{figure}[h]
\vspace{-3mm}
  \centering
  \includegraphics[width=8.5cm, height=5.5cm]{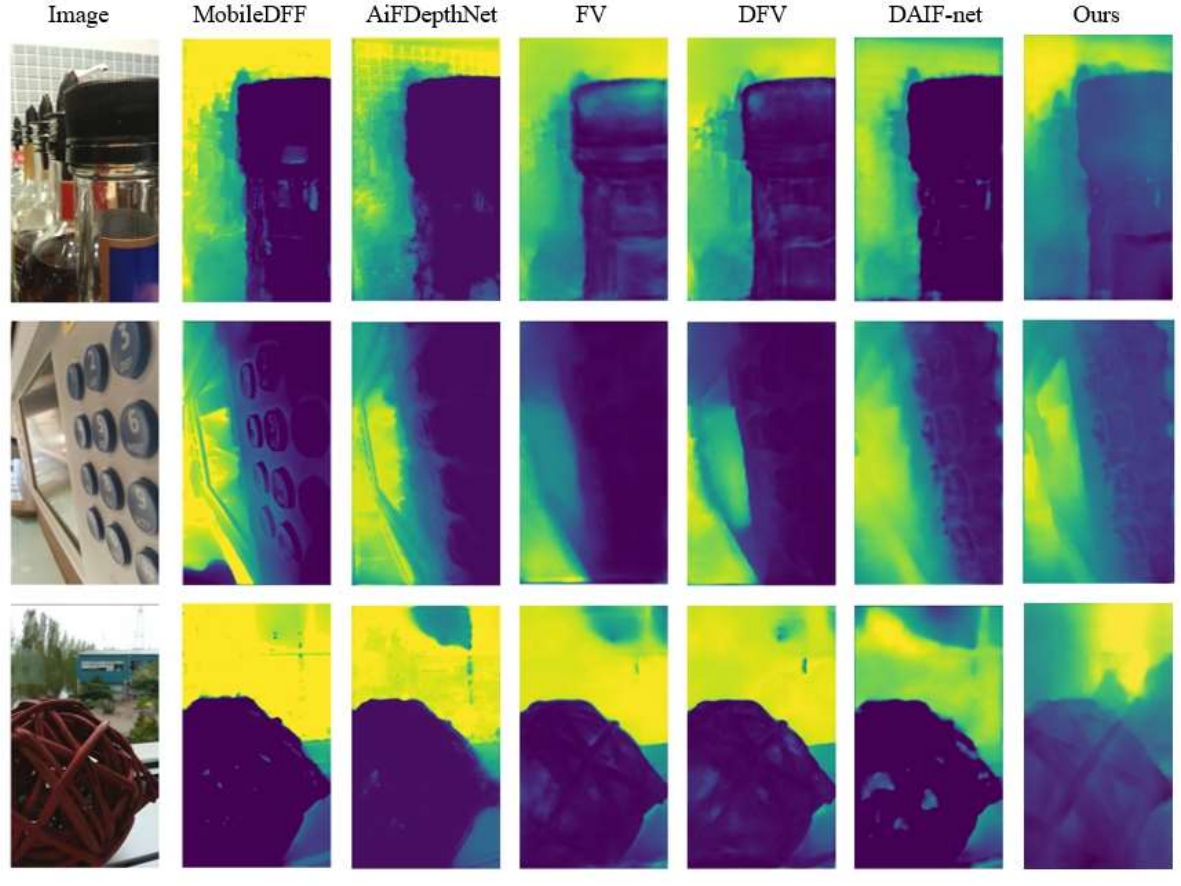}
 \vspace{-5mm}
	\caption{Depth estimation results on MobileDFF dataset. The warmer color indicates a larger depth. We choose DAIFNet\cite{si2023fully}, AiFDepthNet\cite{wang2021bridging}, DFV\cite{yang2022deep},MobileDFF\cite{suwajanakorn2015depth} as a comparsion.}
	\label{fig:mobile}
 \vspace{-2mm}
\end{figure}


\begin{figure}[t]
\begin{center}
\includegraphics[width=8.5cm, height=3.5cm]{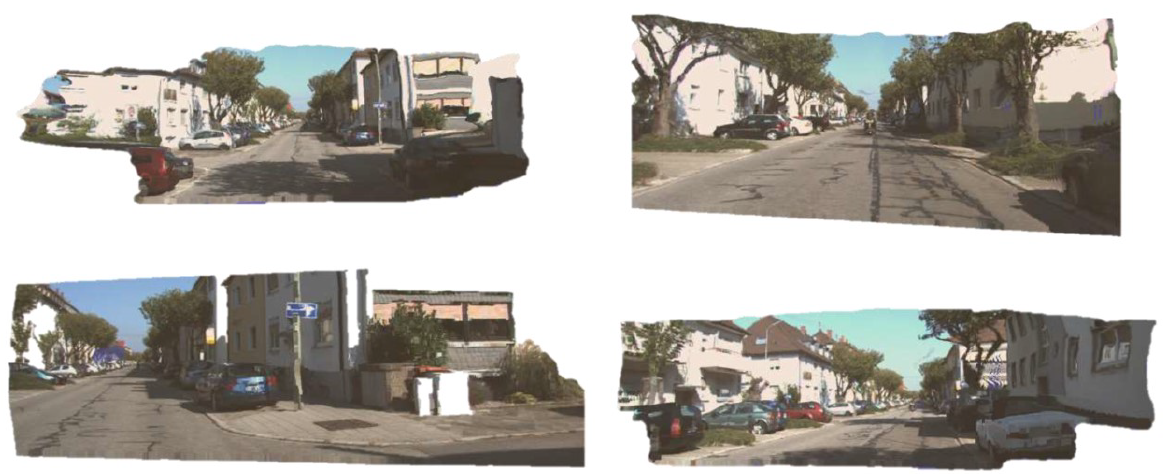}
\end{center}
\vspace{-6mm}
\caption{3D map generated results of KITTI Odometry dataset. Each 3D map is created by merging ten consecutive point clouds..}
\vspace{-3mm}
\label{kittimap}
\end{figure}

\textbf{KITTI Dataset:}  We also evaluated our model on the KITTI dataset. For a sequence of 10 consecutive images, we performed 3D reconstruction by combining the estimated depth with the ground truth pose, as shown in Figure \ref{kittimap}. The point cloud results demonstrate the accuracy of our model .

\textbf{MobileDFF Dataset:} For the MobileDFF dataset, due to the lack of ground truth for defocus maps, we opted for joint training on the FoD500 and NYUv2 datasets before conducting evaluations on the MobileDFF dataset. We selected four methods for comparison: AiFDepthNet\cite{wang2021bridging}, DFF-DFV/FV\cite{yang2022deep}, and DAIFNet\cite{si2023fully}. Based on the visual results depicted in Fig. \ref{fig:mobile}, our method is capable of generating more obvious depth changes even for subtle differences.


\vspace{-2mm}
\subsection{Ablation studies}
\textbf{Loss functions:} we conducted ablation studies on the loss functions of our model to validate the effectiveness of the chosen loss functions. We separately verified the effectiveness of the blurred reconstruction loss and the defocus loss. These experiments were carried out on the NYUv2 dataset, and the quantitative results are presented in Table \ref{Tab:Loss Ablation}. For defocus loss ablation experiments, we change the defocus loss into triplet loss, and do evaluation on NYUv2 dataset. From the table, we can observe that the blurred reconstruction loss and defocus loss can improve the accuracy of our model by 2.2\% and 3.4\%. We can see that with our defocus loss and blurred reconstruction loss, our model can achieve a higher accuracy and lower error rate.

\begin{table}[t]
\begin{center}
\resizebox{0.45\textwidth}{!}{
\begin{tabular}{@{}cc|ccc|cc@{}}
  \toprule
  \multicolumn{2}{c|}{Setting} & \multirow{2}{*}{$\delta_1\uparrow$} & \multirow{2}{*}{$\delta_2\uparrow$} & \multirow{2}{*}{$\delta_3\uparrow$} & \multirow{2}{*}{$RMSE\downarrow$} & \multirow{2}{*}{$AbsRel\downarrow$} \\ \cmidrule(r){1-2}
  \multicolumn{1}{c|}{blurred} & defocus & & & & & \\ \midrule
  \multicolumn{1}{l|}{$\checkmark$} & & 0.943 & 0.982 & 0.992 & 0.254 & 0.079 \\
  \multicolumn{1}{l|}{} & $\checkmark$ & 0.932 & 0.976 & 0.989 & 0.312 & 0.137 \\
  \multicolumn{1}{l|}{$\checkmark$} & $\checkmark$ & 0.964 & 0.998 & 0.999 & 0.201 & 0.026 \\ \bottomrule
\end{tabular}}
\end{center}
\vspace{-3.5mm}
\caption{Ablation experiments for loss functions.}
\label{Tab:Loss Ablation}
\vspace{-11mm}
\end{table}

\begin{table}[h]
\vspace{-2mm}
\begin{center}
\resizebox{0.45\textwidth}{!}{
  \begin{tabular}{@{}ll|lll|ll@{}}
  \toprule
  \multicolumn{2}{l|}{siamese}  & \multirow{2}{*}{$\delta_1\uparrow$} & \multirow{2}{*}{$\delta_2\uparrow$} & \multirow{2}{*}{$\delta_3\uparrow$} & \multirow{2}{*}{$RMSE\downarrow$} & \multirow{2}{*}{$AbsRel\downarrow$} \\ \cmidrule(r){1-2}
  \multicolumn{1}{c|}{wo}           & w            &                                     &                                     &                                     &                                   &                                     \\ \midrule
  \multicolumn{1}{l|}{$\checkmark$} &              & 0.897                               & 0.974                               & 0.985                               & 0.453                             & 0.297                               \\
  \multicolumn{1}{l|}{}             & $\checkmark$ & 0.964                               & 0.998                               & 0.999                               & 0.201                             & 0.026                               \\ \bottomrule
\end{tabular}}
\end{center}
  \vspace{-3.5mm}
  \caption{Ablation experiments for model structure.}
  \label{Tab:MS ablation}
  \vspace{-8mm}
\end{table}

\textbf{Model Structure:} we conduct ablation studies on the structure of the model to prove that the effect of Siamese network in improving the performance of depth estimation. In this ablation study, experiments are performed with the single MPViT model to predict the defocus map without the change of DepthNet. The results are shown in Table \ref{Tab:MS ablation}.

\vspace{-1mm}
\section{Conclusion}
\vspace{-1mm}
This paper introduces a novel framework for depth estimation from defocused images. By incorporating the camera lens model, the network generates images with varying blur levels as input. The framework, built on Siamese networks and 3D Gaussian splatting, is trained in a self-supervised manner. The Siamese network predicts defocus maps and the Circle of Confusion, which are used by the 3D Gaussian splatting model to generate synthetic blurred images. The model parameters are optimized by comparing synthetic and real blurred images. Additionally, the depth corresponding to the blurred images is fed into DepthNet, enabling high-precision depth estimation across various datasets.

\textbf{Acknowledgment}: This work is supported by NSF Grants NO. 2340882, 2334624, 2334246, and 2334690.


\vspace{-2mm}
{\small
\bibliographystyle{ieee_fullname}
\bibliography{egbib}
}

\end{document}